# Transferability of Neural Network Clinical De-identification Systems

[1]Kahyun Lee, Ph.D.*, [2]Nicholas J. Dobbins*, [3]Bridget McInnes, Ph.D., [2]Meliha Yetisgen, Ph.D., [1]Özlem Uzuner, Ph.D., FACMI

[1] George Mason University, Fairfax, VA, USA
[2] University of Washington, Seattle, WA, USA
[3] Virginia Commonwealth University, Richmond, VA, USA

**Abstract**

**Objective**: Neural network de-identification studies have focused on individual datasets. These studies assume the availability of a sufficient amount of human-annotated data to train models that can generalize to corresponding test data. In real-world situations, however, researchers often have limited or no in-house training data. Existing systems and external data can help jump-start de-identification on in-house data; however, the most efficient way of utilizing existing systems and external data is unclear. This article investigates the transferability of a state-of-the-art neural clinical de-identification system, NeuroNER, across a variety of datasets, when it is modified architecturally for domain generalization and when it is trained strategically for domain transfer.

**Methods and Materials:** We conducted a comparative study of the transferability of NeuroNER using four clinical note corpora with multiple note types from two institutions. We modified NeuroNER architecturally to integrate two types of domain generalization approaches. We evaluated each architecture using three training strategies. We measured: transferability from external sources; transferability across note types; the contribution of external source data when in-domain training data are available; and transferability across institutions.

**Results and Conclusions**: Transferability from a single external source gave inconsistent results. Using *additional* external sources consistently yielded an $F_1$-score of approximately 80%. Fine-tuning emerged as a dominant transfer strategy, with or without domain generalization. We also found that external sources were useful even in cases where in-domain training data were available. Transferability across institutions differed by note type and annotation label but resulted in improved performance.

## 1 Introduction

Within biomedical informatics, de-identification refers to the task of removing protected health information (PHI) from clinical text. It aims to unlink health information from patients so that health information can be used without potential infringement of patient privacy. Due to the high cost and limited practicality of manual de-identification [1], automated de-identification systems have been developed. Early efforts at automated de-identification focused on rule-based methods [2,3], using dictionaries of names and common patterns to identify PHI. Subsequent approaches used supervised machine learning methods, treating de-identification as a sequence labeling problem. These approaches included Conditional Random Fields (CRF) [4], Support Vector Machines (SVMs) [5], and Hidden Markov Models (HMMs) [6,7]. The current state-of-the-art de-identification systems, such as our publicly available NeuroNER, are neural network-based [8-11] and have demonstrated promising performance in recent natural language processing (NLP) shared tasks [12,13]. When training and test sets are from the same distribution, these systems detect PHI with high accuracy. However, they fail to generalize to corpora that differ in their distribution of PHI [14].

A handful of studies have tried to mitigate this problem with domain generalization.[1] Lee et al. [15] combined rule-based methods with dictionaries and regular expressions alongside two CRF-based taggers which both incorporated data from the i2b2/UTHealth-2014 dataset using EasyAdapt [16], a domain adaptation technique. Lee et al. [17] examined the generalizability of feature-based machine learning de-identification classifiers trained on an unseen note type (psychiatry). To improve baseline performance, they evaluated several domain generalization techniques including instance pruning, instance weighting, and feature augmentation [16]. They concluded that feature augmentation was the most effective approach in boosting the overall de-identification performance with modest $F_1$-score gains of +0.18 to +0.59 over the next best method. Lee et al. [18] evaluated transfer learning methods on an architecture similar to NeuroNER [8]. They fine-tuned the trained source dataset to the target dataset by measuring which model parameters affected performance the most. The authors noted that transfer learning with only 16% of available in-domain training data achieved comparable performance as in-domain learning with 34% of the training data.

---

[1] Domain adaptation, domain generalization, and transfer learning are closely related concepts. Although there are slight differences in the definitions of these learning concepts, for simplicity we refer to them interchangeably in this paper.

Our de-identification methods focus on finding and removing tokens that correspond to PHI categories defined by the Health Insurance Portability and Accountability Act (HIPAA) Safe Harbor[2] (See Table A.1 in Appendix) and mark the PHI type of each token. We add 'Doctor' and 'Hospital' categories to the HIPAA PHI, since their removal may improve privacy protection [5].

In this article, we evaluate transferability of a state-of-the-art neural network de-identification system, NeuroNER, with and without architectural modifications for domain generalization, and under various training strategies, for identifying narrative text tokens corresponding to PHI. NeuroNER[3] is available online and readily lends itself to architectural modifications. Our architectural modifications include joint learning of domain information with de-identification (referred henceforth as Joint-Domain Learning (JDL)), and adoption of the state-of-the-art domain generalization approach of Common-Specific Decomposition (CSD) [19]. We apply three training strategies, *sequential, fine-tuning, and concurrent,* with NeuroNER, JDL, and CSD, as appropriate. We experiment with datasets that represent different distributions of note types and/or come from different institutions. Different note types represent the clinical sub-language of their corresponding medical specializations [20]. As a result, note types often differ in their linguistic representation, and in their use of PHI. Hence, each dataset can be viewed as being from a separate domain. We define *source* as the dataset to learn from, and *target* as the dataset of application. We define an *external source* as a source that is drawn from a different domain than the target. *In-domain* data refers to data that come from the same domain as the target. Given our neural network architectures, training strategies, and domains, we investigate the following research questions.

(Q1) Can models trained on single domain datasets transfer well?

Individual data sets provide a start to de-identification. However, different individual datasets may transfer differently on new target data. We evaluate each of our datasets for transferability to other domains.

(Q2) Can models become more generalizable by adding more external sources?

With in-domain training, larger training data are generally beneficial in making models more generalizable [18]. However, it is unclear whether using sources from multiple different domains improves generalizability, and by how much, for de-identification. Moreover, different training strategies can be applied when multiple sources are available. For example, one can train a model sequentially using external sources one by one in multi-iteration, or combine sources and train them concurrently in one iteration, or combine external and in-domain data in a fine-tuning fashion.

(Q3) Are domain generalization algorithms helpful for transferability of de-identification systems?

Domain generalization methods attempt to generate common knowledge from external sources with the goal of benefiting from this knowledge when de-identifying as-yet unseen targets. In this paper, we integrate two representative domain generalization methods into NeuroNER architecture.

(Q4) Are external sources useful when sufficient in-domain training data are available?

Neural network de-identification systems using in-domain training data have been demonstrated to perform well in many cases. It is less clear, however, if external sources can be helpful even in cases where sufficient in-domain training data are available. On the one hand, differences from external sources may introduce noise to training parameters and deteriorate overall performance. On the other hand, data from different domains may increase the general understanding of the task and may be useful to reduce the amount of required in-domain training data and/or improve performance.

(Q5) Can models transfer across institutions?

---

[2] https://www.hhs.gov/hipaa/for-professionals/privacy/special-topics/de-identification/index.html
[3] http://neuroner.com/

Transferability across institutions can be different from that within the same institution. For example, although different institutions may use different note authoring conventions, structures, and styles [21], models trained on a certain note type source may transfer well to the same note type target in another institution because they at least share the clinical sub-language of their corresponding medical specializations. Transferability may also be affected by the nature of the PHI contained in the data. Often, the available external de-identification data contain surrogate PHI, i.e., realistic placeholder PHI that obfuscate the originals. Target data may in contrast contain authentic PHI. We investigated transferability across institutions using data from two different institutions. One of our institutions provides surrogate de-identification data whereas the second one contains authentic PHI.

Our work builds on previous findings, adding the following contributions. First, we integrate domain generalization into the architecture of our state-of-the-art de-identification system, NeuroNER. Second, we compare NeuroNER with and without domain generalization in multiple experiments using a variety of dataset combinations and three training strategies. Third, we identify the best performing architecture and training strategy combination and evaluate them on external institutions' data when one of the datasets contains surrogate while the other contains authentic PHI.

## 2 Datasets

We used four datasets for our experiments: i2b2-2006, i2b2/UTHealth-2014, CEGS-NGRID-2016 datasets, and the University of Washington (UW) de-identification dataset. The first three datasets come from multiple clinical contexts and time periods within Partners HealthCare: i2b2-2006 consists of discharge summaries [22], i2b2/UTHealth (University of Texas Health Science Center)-2014 consists of mixed set of note types including discharge summaries, progress reports, and correspondences between medical professionals as well as communication with patients [12,23], and CEGS-NGRID-2016 consists of 1,000 psychiatric intake records [13]. The Partners data contain surrogate PHI that have been generated to be realistic in their context. The details of the surrogate generation process are described in Stubbs et al. [12]. The UW dataset comes from the UW Medical Center and Harborview Medical Center and includes ten note types from 2018-2019. These include admit, discharge, emergency department (ED), nursing, pain management, progress, psychiatry, radiology, social work, and surgery notes. The UW dataset contains authentic PHI.

In order for transferability between domains to be possible, datasets must be consistently labeled. We reviewed the annotation guidelines for each dataset and made adjustments in order to ensure label consistency. We refer to Appendix A.2 for details. Note also that not all HIPAA Safe Harbor PHI categories exist in our data.

An overview of the de-identification datasets after the adjustment is shown in Table 1. The i2b2-2006 and CEGS-NGRID-2016 datasets consist of only one note type, whereas the i2b2/UTHealth-2014 dataset is a mixture of multiple note types, and the UW dataset contains ten different note types. Each note within the Partners HealthCare datasets contained 20.6 PHI labels on average, while the UW dataset contains 62.5 PHI labels on average. In terms of proportions of PHI per token, the UW dataset was the densest with an average of 30 PHI labels per note. In contrast, the CEGS-NGRID-2016 dataset was the sparsest, with only one of every 110 tokens being PHI.

|  | **i2b2-2006** | **i2b2/UTHealth-2014** | **CEGS-NGRID-2016** | **UW** |
| --- | --- | --- | --- | --- |
| Institution | Partners HealthCare | Partners HealthCare | Partners HealthCare | UW Medicine |
| Record types | Discharge notes | Diabetic longitudinal notes | Psychiatric intake notes | admit, discharge, ED, nursing, pain management, progress, psychiatry, radiology, social work, and surgery notes |
| # of records (training/test) | 889 (669/220) | 1,304 (790/514) | 1,000 (600/400) | 1,000 (800/200) |
| # of tokens | 586,186 | 1,066,224 | 2,305,206 | 1,900,402 |
| Average # of tokens per record | 659 | 818 | 2,305 | 1,900 |
| # of PHI | 19,498 | 25,260 | 21,121 | 44,701 |
| Average # of PHI per record | 22 | 19 | 21 | 44 |

| Average # of token per PHI token | 30 | 42 | 110 | 42 |

**Table 1.** Overview of the de-identification datasets

## 3 Methods

Our methods identify narrative text tokens that correspond to PHI and specify the type of PHI for each token. Table A.3. in the appendix shows the distribution of the PHI types covered in our datasets. As can be seen, the PHI distributions vary across datasets but represent the realities that would be faced by any institution in trying to jumpstart their de-identification efforts with external data.

### 3.1. Baseline De-identification approach – NeuroNER

To measure the transferability of neural network de-identification models, we used our NeuroNER [8], a representative, publicly available, state-of-the-art bidirectional Long Short-Term Memory with Conditional Random Fields (biLSTM-CRF) system which utilizes character and word embeddings [24, 25] as features. The system first splits an input string into sentences. Each character in a sentence is assigned to a unique character embedding, which is then passed through a character-level biLSTM layer. The output from that layer is concatenated with pretrained word embeddings to create character-enhanced token embeddings which are fed into a token-level biLSTM layer. The system performs multi-label prediction of PHI by delivering token-level biLSTM output into a fully-connected layer and produces probabilities for individual PHI labels for each token in the sentence. The final PHI label for each token is determined by a CRF that maximizes the likelihood of the PHI label sequence for a sentence.

### 3.2. Domain generalization

We incorporated two domain generalization algorithms into NeuroNER.

*Common-Specific Decomposition (CSD):* Decomposition is one of the major themes in domain generalization [19,26-28]. In general decomposition training, parameters in a neural network layer are considered as the sum of their common parameters (i.e., general parameters across multi-domains) and domain-specific parameters (i.e., specifically related to one domain). Piratla et al. [19] suggest an efficient variation of this technique using common-specific low-rank decomposition (CSD). Assuming there are common features which are consistent across domains, they decompose the last layer of a neural network system into common parameters and domain-specific parameters. The domain-specific parameters are calculated as being low-rank and orthogonal to the common parameters. The losses from both the common and domain-specific parameters are backpropagated during training. When making predictions, however, domain-specific parameters are dropped, and only the common parameters are used. In this way, the effect of domain differences is minimized. The overview of the CSD implementation on NeuroNER is shown in the left-side of Figure 1 where CSD was integrated into the penultimate layer of the original NeuroNER.

*Joint Domain Learning (JDL):* JDL builds on the principles of joint learning [28]. It treats de-identification and domain prediction as two tasks that can be learned jointly and that can inform each other. In JDL a separate fully connected layer enables NeuroNER to jointly learn both a sequence of labels and a domain per each sentence. As shown in the right-side of Figure 1, the Fully-Connected Layer 2 in the shaded area is used only to learn and predict the domain of the input sentence, while the Fully-Connected Layer 1 is used for label prediction. During training, losses from the label prediction and domain prediction layers are combined and backpropagated to update parameters in earlier layers. The two losses can be combined either equally or disproportionately. We experimented with various ratios and found the best performance with 85% of loss from label prediction and 15% of loss from domain prediction. For testing, the model predicts based on Fully-Connected Layer 1 followed by CRF and does not use the Fully-Connected Layer 2.

## 4 Experiments

### 4.1 Experimental Setting

Our experiments evaluated three architectures, NeuroNER, JDL, and CSD, using different training strategies. We defined three training strategies, *sequential*, *fine-tuning,* and *concurrent* as follows.

*Sequential*: This approach starts with training using a single external dataset to create pretrained weights which can be further refined on the *external* datasets in an iterative fashion. We experimented with different sequential ordering of domains; however, only the highest performing sequence of each combination was used for the follow up experiments. Note that sequential training starts with learning from a single external dataset and can only be used for training NeuroNER. JDL and CSD by default require two or more domains to learn from and are therefore excluded from sequential training experiments.

*Fine-tuning*: A special form of sequential training where the in-domain training data is the final sequentially-trained dataset.

*Concurrent*: Concurrent training combines multiple external sources and trains in a single iteration.

Experiments were conducted to respond to the previously presented research questions:

1. To assess whether models trained on single domain datasets transfer well (Q1), we train on individual i2b2-2006, i2b2/UTHealth-2014, and CEGS-NGRID-2016 datasets and evaluate against each of their respective test sets (e.g., training and testing on i2b2-2006) as well as against each other's test sets (e.g., training i2b2-2006 and testing on i2b2/UTHealth-2014) using NeuroNER to establish baselines.
2. To check whether models become more generalizable by adding more external sources (Q2) and whether domain generalization architectures are helpful for transferability (Q3), we train with each of the three architectures with different training strategies using only external sources (e.g., training a mix of i2b2-2006 and i2b2/UTHealth-2014, and testing on CEGS-NGRID-2016).
3. To evaluate whether external sources are useful when sufficient in-domain training data are available (Q4), we repeat the experiments above but include in-domain training data (e.g., training a mix of i2b2-2006, i2b2/UTHealth-2014, CEGS-NGRID-2016, and testing on CEGS-NGRID-2016).
4. To measure transferability across institutions (Q5), we apply the best-performing architecture and training strategy combination across institutions, using one institution as the source and the other as the target.

### 4.2 Training

We tune our hyperparameters on the entire Partners HealthCare training set. The resulting hyperparameters were as follows: character embeddings dimension = 25; token embeddings dimension = 100; hidden layer dimension for LSTM over character = 25; hidden layer dimension for LSTM over concatenation of {character LSTM output, token embeddings} = 100; dropout = 0.5; optimizer = SGD; learning rate = 0.005; maximum number of epochs = 100; patience = 10. Note that it would also have been valid to tune the parameters to individual datasets. However, our experiments showed that the above parameters worked well for individual data sets as well with very few exceptions where dataset-specific parameters would have affected performance. For the sake of simplicity, we chose to adopt the above parameters for all experiments.

### 4.3 Evaluation

Among various evaluation metrics for de-identification [12], we used PHI (entity-based) $F_1$-scoring which is common in named entity recognition evaluation. In our case, PHI represent entities, with $F_1$ calculated as 2*(precision*recall) / (precision + recall), where precision is defined as true positives over the sum of true positives and false positives, and recall is defined as true positives over the sum of true positives and false negatives. The entity-based $F_1$-score requires exact PHI spans to be found, and counts partial predictions as false. For example, the entity "Jane Doe" is composed of 2 tokens, both of which would have to be marked as 'Patient' PHI in order to count as correct. Partial labels that include only "Jane" or "Doe" are considered incorrect.

### 4.4 Results

Table 2 presents the performance of NeuroNER on the individual i2b2 datasets (in response to Q1). Using in-domain training data, the system achieved $F_1$-scores of 96.9, 94.0, and 92.5 on the i2b2-2006, i2b2/UTHealth-2014, and CEGS-NGRID-2016 test sets, respectively. The $F_1$-scores of the models trained on a single external source were notably poorer, ranging from 58.2 to 82.3. The model trained on the i2b2-2006 dataset was the least generalizable and achieved $F_1$-scores of 64.4 and 58.2 on the i2b2/UTHealth-2014 and CEGS-NGRID-2016 test data. The model trained using i2b2/UTHealth-2014 dataset was the most generalizable, achieving $F_1$-scores of 75.2 and 82.3 on the i2b2-2006 and CEGS-NGRID-2016 test sets. This may be because i2b2/UTHealth-2014 dataset contains various note types whereas i2b2-2006 and CEGS-NGRID-2016 datasets consist of a single note type.

| Architecture | Source | Target | | |
|---|---|---|---|---|
| | | i2b2-2006 | i2b2/UTHealth-2014 | CEGS-NGRID-2016 |
| NeuroNER | 2006 | **97.8/96.1/96.9** | 73.9/57.0/64.4 | 68.6/50.5/58.2 |
| | 2014 | 77.4/73.2/75.2 | **94.3/93.8/94.0** | 84.0/80.1/82.3 |
| | 2016 | 68.1/60.9/64.3 | 75.4/71.6/73.5 | **93.4/91.6/92.5** |

**Table 2.** Baseline results. Performance when trained on individual datasets (%, P/R/$F_1$). Bold indicates in-domain evaluation.

To check whether models become more generalizable by adding more external sources (Q2) and whether domain generalization is helpful for transferability (Q3), we experimented using two external sources with the NeuroNER, CSD, and JDL architectures. For NeuroNER, sequential and concurrent training strategies were used. Because CSD and JDL utilize two or more external sources without any consideration for sequence, their experiments are considered concurrent. Results are shown in Table 3. Compared to Table 1, the results were largely consistent for all targets and higher overall, with NeuroNER using a concurrent training strategy achieving the best results. CSD and JDL performed higher on the i2b2-2006 dataset, but generally performed worse than NeuroNER for all training strategies. JDL performed higher than CSD in all cases.

| Architecture | Training Strategy | Source | Target | | |
|---|---|---|---|---|---|
| | | | i2b2-2006 | i2b2/UTHealth-2014 | CEGS-NGRID-2016 |
| NeuroNER | Sequential | 2014→2016 | 77.4/73.2/75.2 | - | - |
| | | 2016→2014 | 78.2/79.4/78.8 | - | - |
| | | 2006→2016 | - | **80.2/77.4/78.8** | - |
| | | 2016→2006 | - | 75.4/71.6/73.5 | - |
| | | 2006→2014 | - | - | **85.5/82.3/83.9** |
| | | 2014→2006 | - | - | 84.0/80.1/82.3 |
| | Concurrent | 2014+2016* | 77.6/81.5/79.5 | - | - |
| | | 2006+2016 | - | 80.1/77.1/78.6 | - |
| | | 2006+2014 | - | - | 82.7/80.8/81.7 |
| CSD | Concurrent | 2014+2016 | 82.8/78.6/80.6 | - | - |
| | | 2006+2016 | - | 82.1/69.2/75.1 | - |
| | | 2006+2014 | - | - | 81.4/65.5/72.6 |
| JDL | Concurrent | 2014+2016 | **81.8/81.7/81.7** | - | - |
| | | 2006+2016 | - | 80.6/71.8/76.0 | - |
| | | 2006+2014 | - | - | 79.5/68.8/73.8 |

**Table 3.** Performance of trained models without in-domain data (%, P/R/$F_1$). Bold indicates best performance for each target test set. → indicates sequential learning. + indicates concurrent learning. Results are reported only on the target test set. i.e., i2b2-2006→i2b2/UTHealth-2014 sequentially trains on i2b2-2006 training set, continues training on the i2b2/UTHealth-2014 training set, and applies the resulting model to CEGS-NGRID-2016 test set.

To evaluate whether external sources are useful when sufficient in-domain training data are available (Q4), we repeated the experiments above but included in-domain training data for fine-tuning. Results are shown in Table 4. Fine-tuning experiments with concurrent training and NeuroNER architecture showed improvement over results in Table 1 using only in-domain training data, and a significantly higher performance than Table 3 external source-only experiments. As in Table 3, $F_1$-scores for fine-tuned CSD and JDL architectures performed at or slightly higher than NeuroNER fine-tuned on the i2b2-2006 test dataset, but worse on the i2b2/UTHealth-2014 and CEGS-NGRID-2016 datasets. As in Table 3, JDL performed higher than CSD in all cases.

| Architecture | Training Strategy | Source | Target | | |
|---|---|---|---|---|---|
| | | | i2b2-2006 | i2b2/UTHealth-2014 | CEGS-NGRID-2016 |
| NeuroNER | Fine-tuning from Sequential | 2016→2014→2006 | 97.8/96.5/97.2 | - | - |
| | | 2006→2016→2014 | - | 95.8/94.9/95.3 | - |
| | | 2006→2014→2016 | - | - | 93.8/92.0/92.9 |
| | | 2014+2016→2006 | 97.9/96.8/97.3 | - | - |
|  | Fine-tuning from Concurrent | 2006+2016→2014 | - | **95.9/95.2/95.5** | - |
|  |  | 2006+2014→2016 | - | - | **93.9/92.6/93.2** |
|  | Concurrent | 2006+2014+2016 | 96.6/97.2/96.8 | 95.6/94.9/95.3 | 94.1/92.0/93.0 |
| CSD | Fine-tuning from Concurrent | 2014+2016→2006 | 98.1/96.4/97.2 | - | - |
|  |  | 2006+2016→2014 | - | 94.9/91.0/92.9 | - |
|  |  | 2006+2014→2016 | - | - | 92.1/86.3/89.1 |
|  | Concurrent | 2006+2014+2016 | 96.3/95.9/96.1 | 94.7/91.4/93.0 | 92.6/84.1/88.1 |
| JDL | Fine-tuning from Concurrent | 2014+2016→2006 | **98.0/97.0/97.5** | - | - |
|  |  | 2006+2016→2014 | - | 95.6/91.2/93.4 | - |
|  |  | 2006+2014→2016 | - | - | 90.9/87.6/89.2 |
|  | Concurrent | 2006+2014+2016 | 96.5/95.6/96.1 | 95.0/91.2/93.1 | 91.3/85.3/88.2 |

**Table 4.** Performance of trained models with in-domain datasets (%, P/R/$F_1$). Bold indicates best performance for each target test set by $F_1$-score. → indicates sequential learning. + indicates concurrent learning. Results are only reported on the target test set.

Finally, we explored transferability across institutions (Q5) by first training a model using the training data of the combined Partners HealthCare datasets (i2b2-2006, i2b2/UTHealth-2014, CEGS-NGRID-2016) and applying the model to UW test data. We then reversed the direction on these experiments and applied the model from UW to Partners data. Results are shown in Table 5. As previously mentioned, Partners data contain surrogate PHI that are generated to be realistic placeholders while the UW data contains authentic PHI.

| Architecture | Training Strategy | Source | Target | Performance |
|---|---|---|---|---|
| NeuroNER | In-domain | UW | UW | 90.5/91.8/91.1 |
|  | Out-domain | 2006+2014+2016 | UW | 77.7/79.2/78.5 |
|  |  |  |  | 78.6/77.7/78.1 |
|  | Concurrent | 2006+2014+2016+UW | UW | 90.1/91.9/91.0 |
|  | Fine-tuning | 2006+2014+2016→UW | UW | **91.1/92.3/91.7** |
|  | In-domain | 2006+2014+2016 | 2006+2014+2016 | 91.1/86.3/88.6 |
|  | Out-domain | UW | 2006+2014+2016 | 85.3/65.2/73.9 |
|  | Concurrent | UW+ 2006+2014+2016 | 2006+2014+2016 | 92.2/85.9/89.0 |
|  | Fine-tuning | UW →2006+2014+2016 | 2006+2014+2016 | **92.1/86.7/89.3** |

**Table 5.** Transferability across institutions (%, P/R/$F_1$). Best performance by $F_1$-score is in bold.

$F_1$ performance of in-domain UW training was 91.1. We achieved the best performance by fine-tuning a pretrained model using data from the UW training set, which improved performance to 91.7 (+0.6) on the UW test set. We calculated statistical significance between UW in-domain (91.1) and fine-tuned from Partners HealthCare (91.7) test results using approximate randomization [29] and 10,000 shuffles. The differences in $F_1$-score were statistically significant ($p \leq 0.05$). In the reverse direction, the combined Partners datasets achieved the best performance also using fine-tuning with the UW training dataset with an $F_1$-score of 89.3 versus a concurrent training score of 89.0. Differences between Partners HealthCare in-domain (89.0) and fine-tuned from UW (89.3) were also statistically significant.

The sensitivity for the transfer across institutions varied per note type and label. To demonstrate performance differences across note types and labels, more granular examination of results can be found in the Appendix B, Tables B.1, B.2, and Figure B.1. Table B.1 shows UW test set performance by note type using each i2b2 dataset for training, with Admit, Nursing, and Surgery notes performing notably well. Figure B.1 shows test results for UW Discharge and Psychiatry notes trained on each i2b2 dataset broken down by precision, recall, and $F_1$-score. Table B.2 compares UW test results by PHI type with and without fine-tuning, with 'ID' and 'Phone' types showing the most notable improvement. Given that our experiments found the i2b2/UTHealth-2014 dataset to be the most generalizable, we also conducted further experiments mirroring those of Table 5 but using only the UW and i2b2/UTHealth-2014 dataset. Results for these are shown in Table B.3 and show gains from even a single external data source when the source contains multiple note types.

## 5 Discussion

Overall, in response to whether models trained on single domain datasets can transfer well (Q1), we found that models trained on single domain datasets showed low and inconsistent performance when applied to target datasets. Among the Partners HealthCare datasets, only the i2b2/UTHealth-2014 dataset includes multiple note types and achieved the best performance among external training sets on i2b2-2006 and CEGS-NGRID-2016 test sets in our baseline experiments using NeuroNER. This suggests that single domain datasets with a greater variety of note types can be expected to generalize better than those with only a single note type.

In terms of whether combining data from multiple source domains improves generalizability to target domains (Q2), as shown in Table 3, we found this to be true but only under certain conditions. When using sequential training, we found the order of training sequence to be critical. When trained in certain sequences, sequential training of NeuroNER yielded +3.6, +5.3 and +1.6 $F_1$-score improvements over the baseline results on all Partners HealthCare datasets. When trained in opposite order from the best-scoring sequences, however, the models failed to improve at all and performance deteriorated as training progressed. We suspect the similarity of the external and in-domain corpora play a role in the success of transfer, but defining similarity is beyond the scope of this article and is left for future work. This suggests careful examination of the relatedness between domains is necessary when choosing the sequence of model training.

This lesson also was true for domain generalization algorithms (Q3). These algorithms generated transferable models. They even provided a stronger model for transfer to the i2b2-2006 dataset, but generally performed worse than concurrent and best-performing sequential training of NeuroNER. Domain generalization approaches benefited from fine-tuning, as did other concurrent training strategies. Perhaps unsurprisingly, the domain generalization architectures were most effective when domain differences among sources were not significant. When the i2b2-2006 dataset (which was the least generalizable in baseline) was used for training, domain generalization approaches did not generalize as well. However, it should be noted that domain generalization approaches thrive when given larger number of training domains. For example, CSD used data from 25~76 domains when it achieved the state-of-the-art performance [19]. Therefore, domain generalization performance may be different if more sources from different note types were available. In all cases of our experiments, JDL performed higher than CSD.

We also observed that even in cases when sufficient in-domain data are available, a model pretrained using external sources provides advantages (Q4). To test performance gains from external data, in the presence of in-domain data, we used two of the Partners corpora as external sources while using the third one as in-domain fine-tuning data and tested performance on the corresponding in-domain test data. We increased the amount of available in-domain fine-tuning data at each iteration, we compared each iteration to the baseline of training on the same amount of in-domain training data and testing on the corresponding in-domain test data. In Figure 2, the dotted lines represent the baseline, and the solid lines represent the model fine-tuned from external data. The solid lines not only ascend faster than the dotted lines, but also ultimately reach a higher level of performance. Repeating the same experiment on UW data, Figure 3 shows performance changes on the UW test set when fine-tuned from combined Partners data using increasing UW training notes in increments of 100. UW note types with comparatively less PHI, such as Surgery, Radiology, and Nursing (see Appendix Table A.4) and Admit notes (which we are hypothesize are structurally similar to Partners notes) perform reasonably well using only 100 notes (10 of each note type), exceeding $F_1$-scores of 95%. Other UW note types which have greater amounts of PHI or tend to be structurally more different than Partners notes perform worse using 100 notes, but generally exceed an $F_1$-score of 90% using 200 or so notes. The exception to this is Social Work notes, which exceeds an $F_1$-score of 80% only when using 800 total training notes.

Based on these findings, if the aim is to achieve an $F_1$-score of 90%, a researcher must annotate about 300 in-domain notes regardless of note type if no out-domain data are present. However, using a pretrained model, our findings suggest that only 50-150 in-domain annotated notes would be needed, depending on the note type. These results are largely consistent with the findings of [17,18].

Regarding transferability across institutions (Q5), we found that models were able to transfer, but as noted earlier sensitivity varied by note type and label. While we expected the i2b2-2006 and CEGS-NGRID-2016 results to show the best performance on UW notes of the same type (discharge and psychiatry), as Appendix 2 shows the i2b2/UTHealth-2014 dataset generalized well (and often better) for both note types, suggesting that certain corpora of mixed note types (such as i2b2/UTHealth-2014) may generalize better than those of single note types.

Additionally, we found that domain transfer performance varied significantly by PHI type (see results in Appendix B). In general, 'Date' types are the easiest to be transferred, as in most cases, dates are transcribed in only a handful of number formats and with a relatively limited vocabulary compared to other PHI. The second most transferable PHI type was 'Doctor', which we assume to be due to common usage of honorifics and degrees such as 'Dr.' and 'M.D.', or indicators such as 'dictated by' and 'signed by' which are used relatively consistently across institutions. Other PHI types, particularly 'ID', showed generally poor performance in transferability. 'Age' types, as limited to only cases with patients over 90 years and thus rarely present in the training data, proved to predictably show the poorest performance.

# 6 Conclusion

In this study, we investigated the transferability of neural network de-identification methods from the practical perspective of researchers having a limitation on preparing enough in-domain annotated data. We found de-identification model transfers from a single external source to be limited. However, a more generalizable model can be produced using two or more external sources, which consistently achieved an $F_1$-score of approximately 80%. We also found that domain generalization techniques generated transferable models but could also benefit from fine-tuning . Additionally, external sources were useful even in cases where in-domain annotated data exist by reducing the required amount of in-domain annotation data and/or further improving model performance. The models could be transferred across institutions, however, the sensitivity for the transfer differed between note types and labels. When transferring from another institution, data of mixed note types performed higher than single note type data. External sources from different institutions were still useful to further improve performance in cases where in-domain training data were available. We expect these findings will be useful in allowing de-identification models to be more easily leveraged across institutions.

## Acknowledgments

This study was supported in part by the National Library of Medicine under Award Number R15LM013209 and by the National Center for Advancing Translational Sciences of National Institutes of Health under Award Number UL1TR002319. Experiments on the UW data were run on computational resources generously provided by the UW Department of Radiology. We gratefully acknowledge the work of David Wayne (UW) in the annotation of clinical notes.

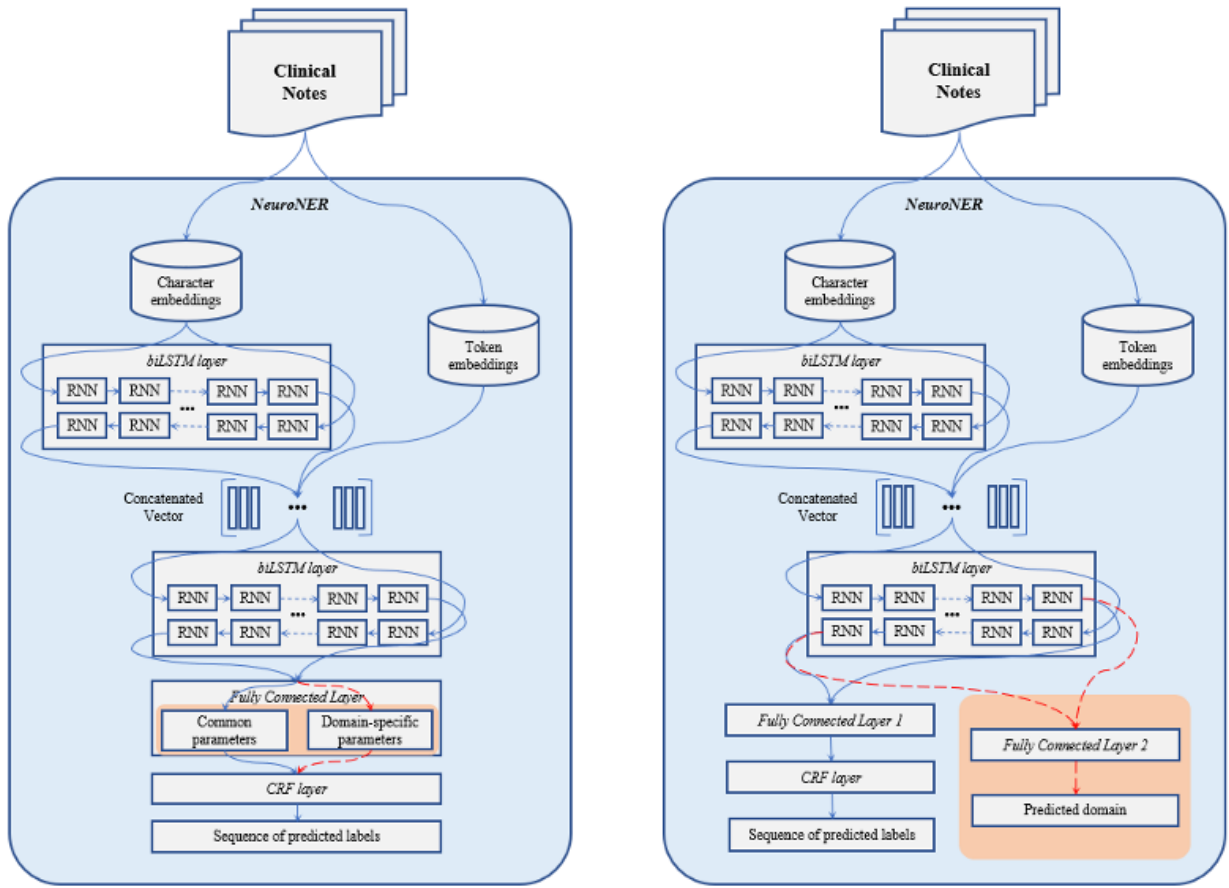

**Figure 1.** Overview of Common-Specific Decomposition (left) and Join Domain Learning (right). Structures for domain generalization are shaded. The dotted lines are used only in training.

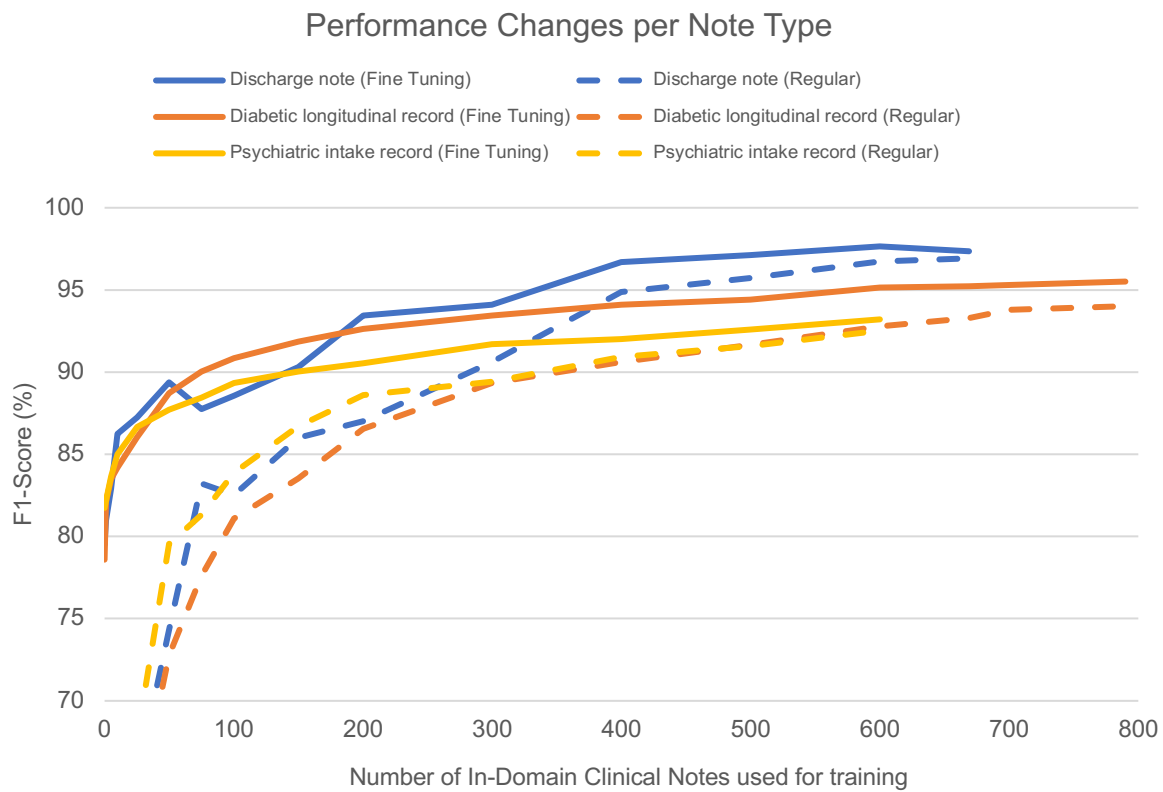

**Figure 2**. Learning curve per in-domain training size using i2b2-2006, i2b2/UTHealth-2014, and CEGS-NGRID-2016 datasets (%, $F_1$)

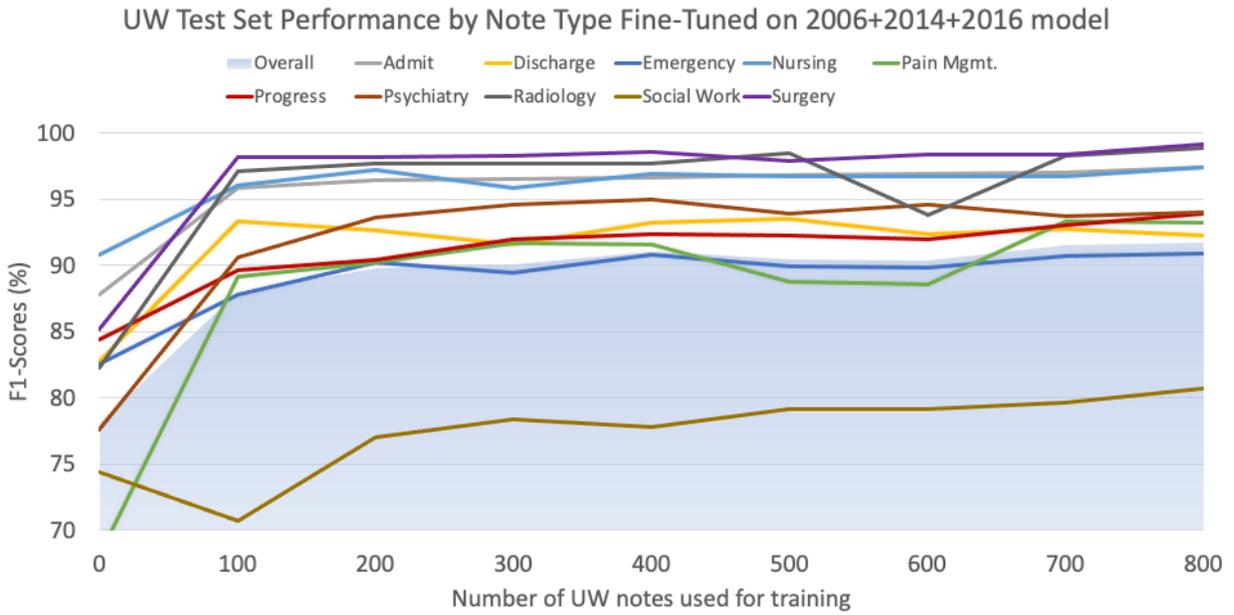

**Figure 3** Learning curve per in-domain UW training size using fine-tuning on a 2006+2014+2016 pretrained model. Each iteration incrementally increased the number UW notes for training, with each note type representing 10% of training data (e.g., for 100 total UW training notes, 10 from each note type were randomly chosen). Performance is measured by $F_1$ scores using the same 200 UW test notes at each stage.

# Appendix

The following identifiers of the individual or of relatives, employers, or household members of the individual:
(1) Names
(2) All geographic subdivisions smaller than a State, including street address, city, county, precinct, zip code, and their equivalent geocodes, except for the initial three digits of a zip code if, according to the current publicly available data from the Bureau of the Census: 1. The geographic unit formed by combining all zip codes with the same three initial digits contains more than 20,000 people; and 2. The initial three digits of a zip code for all such geographic units containing 20,000 or fewer people is changed to 000.
(3) All elements of dates (except year) for dates directly related to an individual, including birth date, admission date, discharge date, date of death; and all ages over 89 and all elements of dates (including year) indicative of such age, except that such ages and elements may be aggregated into a single category of age 90 or older
(4) Telephone numbers
(5) Fax numbers
(6) Electronic mail addresses
(7) Social security numbers
(8) Medical record numbers
(9) Health plan beneficiary numbers
(10) Account numbers
(11) Certificate/license numbers
(12) Vehicle identifiers and serial numbers, including license plate numbers
(13) Device identifiers and serial numbers
(14) Web Universal Resource Locators (URLs)
(15) Internet Protocol (IP) address numbers
(16) Biometric identifiers, including finger and voice prints
(17) Full face photographic images and any comparable images
(18) Any other unique identifying number, characteristic, or code

**Table A.1** HIPAA Safe Harbor PHI Categories

## A. Dataset

PHI for each dataset followed guidance from HIPAA documentation. To align the datasets, we implemented the following changes:

- The i2b2 datasets vary in interpretation of whether 'Date' PHI encompass years, or only months and days; i2b2-2006 excludes years, while subsequent Partners HealthCare datasets include them. We removed all year PHI components from 'Date' annotations from the i2b2/UTHealth-2014 and CEGS-NGRID-2016 datasets in order to align Date annotations with the lowest common denominator dataset, i2b2-2006. This was done using regular expressions followed by manual inspection and validation.
- 'Email', 'URL', 'Organization' and 'Profession' type PHI were not labeled in the i2b2-2006 dataset; therefore, we do not utilize these labels from the i2b2/UTHealth-2014 and CEGS-NGRID-2016 datasets.
- In the i2b2-2006 i2b2 shared task, 'IDs' were defined as "any combination of numbers, letters, and special characters identifying medical records, patients, doctors, or hospitals", which we interpreted to correspond to the 'MedicalRecord', 'Device', 'HealthPlan', 'License', 'BioID', 'IDNUM', and 'Username' labels present in i2b2/UTHealth-2014 and CEGS-NGRID-2016 shared task. We therefore mapped these labels to a single 'ID' label.
- All ages were annotated in i2b2/UTHealth-2014 and CEGS-NGRID-2016 dataset, but only ages over 89 were labeled in i2b2-2006. We therefore removed any ages labeled which were under age 90 from i2b2/UTHealth-2014 and CEGS-NGRID-2016 datasets.

The UW dataset was annotated according to i2b2/UTHealth-2014 and CEGS-NGRID-2016 de-identification guidelines. We therefore followed the same steps outlined above and similarly aligned the UW annotations with the i2b2-2006 data. Our adjustments are summarized in Table A.2.

| i2b2-2006 | i2b2/UTHealth-2014 / CEGS-NGRID-2016 / UW | Adjusted Labels |
| --- | --- | --- |
| Patient | Patient | Patient |
| Doctor | Doctor | Doctor |
| Hospital | Hospital | Hospital |
| ID | MedicalRecord, Device, HealthPlan, License, BioID, IDNUM, Username | ID |
| Date (excluding year) | Date (including year) | Date (excluding year) |
| Location | Street, City, State, Zip, Country, Location, Location-Other | Location |
| Phone | Phone, Fax | Phone |
| Age (over 89) | Age (all) | Age (over 89) |
| - | Email, URL, Organization, Profession | - |

**Table A.2.** PHI type mapping between i2b2-2006 and other datasets.

| PHI TYPE | I2b2-2006 | I2b2/UTHealth-2014 | CEGS-NGRID-2016 | UW |
| --- | --- | --- | --- | --- |
| Age | 16 | 14 | 22 | 6 |
| Date | 7,098 | 11,484 | 6,576 | 23,544 |
| Doctor | 3,751 | 4,797 | 3,963 | 6,197 |
| Hospital | 2,400 | 2,312 | 3,523 | 3,851 |
| ID | 4,809 | 1,862 | 102 | 1,381 |
| Location | 263 | 2,062 | 4,563 | 5,299 |
| Patient | 929 | 2,195 | 2,107 | 3,094 |
| Phone | 232 | 534 | 265 | 1,329 |
| Total | 19,498 | 25,260 | 21,121 | 44,701 |

**Table A.3.** PHI counts by type for each de-identification dataset.

| UW Note Type | AGE | DATE | DOCTOR | HOSPITAL | ID | LOCATION | PATIENT | PHONE |
| --- | --- | --- | --- | --- | --- | --- | --- | --- |
| Admit | 0 | 3,695 | 312 | 295 | 17 | 196 | 297 | 119 |
| Discharge | 0 | 2,874 | 784 | 813 | 111 | 682 | 300 | 423 |
| Emergency | 0 | 2,043 | 1,271 | 181 | 294 | 83 | 114 | 112 |
| Nursing | 0 | 8,62 | 36 | 192 | 12 | 43 | 10 | 2 |
| Pain Mgmt. | 0 | 4,070 | 1,477 | 466 | 709 | 2,637 | 248 | 83 |
| Progress | 1 | 3,539 | 559 | 470 | 34 | 458 | 359 | 71 |
| Psychiatry | 4 | 2,615 | 675 | 514 | 15 | 300 | 1,233 | 69 |
| Radiology | 0 | 416 | 170 | 32 | 10 | 106 | 7 | 24 |
| Social Work | 1 | 638 | 829 | 841 | 129 | 768 | 506 | 410 |
| Surgery | 0 | 2,792 | 84 | 47 | 50 | 26 | 20 | 16 |
| Total | 6 | 23,544 | 6,197 | 3,851 | 1,381 | 5,299 | 3,094 | 1,329 |

**Table A.4.** PHI counts for each note type in the UW dataset.

B. Sensitivity for the transfer across institutions per note type and label

Table B.1 shows UW test set results by note type for each individual Partners HealthCare training sets using NeuroNER architecture, and Figure B.1 shows results by label on UW discharge and psychiatry notes. Table B.2 presents transferability results for the UW test dataset without pretraining and with Partners HealthCare pretrained performance broken down by entities.

|  | | Source | | |
|---|---|---|---|---|
|  | | i2b2-2006 | i2b2/UTHealth-2014 | CEGS-NGRID-2016 |
| **Target** (UW) | Admit | 88.9/81.0/84.4 | **90.2/85.8/88.0** | 85.6/83.2/84.4 |
|  | Discharge | 86.2/73.7/79.4 | 85.6/75.3/80.1 | **88.3/73.8/80.4** |
|  | Emergency | **84.8/72.5/78.1** | 83.8/72.5/77.7 | 86.6/65.3/74.4 |
|  | Nursing | 85.9/78.7/82.1 | **94.1/87.1/90.4** | 79.7/83.6/81.6 |
|  | Pain Mgmt. | 63.2/46.6/53.6 | **72.2/61.6/66.5** | 77.5/53.9/63.6 |
|  | Progress | 87.4/71.9/78.9 | 83.4/82.8/83.1 | **86.5/81.1/83.7** |
|  | Psychiatry | 84.8/44.4/58.3 | **90.7/58.3/71.0** | 86.6/56.6/68.5 |
|  | Radiology | 80.7/79.9/80.3 | **82.5/87.6/85.0** | 81.5/84.0/82.7 |
|  | Social Work | 70.5/49.2/57.9 | **75.8/66.0/70.5** | 73.4/66.4/69.7 |
|  | Surgery | 83.6/96.7/89.7 | **92.4/96.2/94.3** | 80.0/91.9/85.5 |

**Table B.1.** Performance by UW note type and Partners HealthCare training set using NeuroNER architecture (%, P/R/$F_1$). The highest scores for each note type by $F_1$-score are highlighted in bold. The i2b2/UTHealth-2014 achieved the best performance in 7 of 10 note types. The best $F_1$-scores for Nursing, Admit, and Surgery notes were notably high (90.4%, 88.0%, and 94.3%) using this source training set, and are comparable to $F_1$-scores on the i2b2/UTHealth-2014 target set shown in Table 1. We hypothesize that these note types are perhaps more structurally similar between institutions but leave a more robust examination of this to future work.

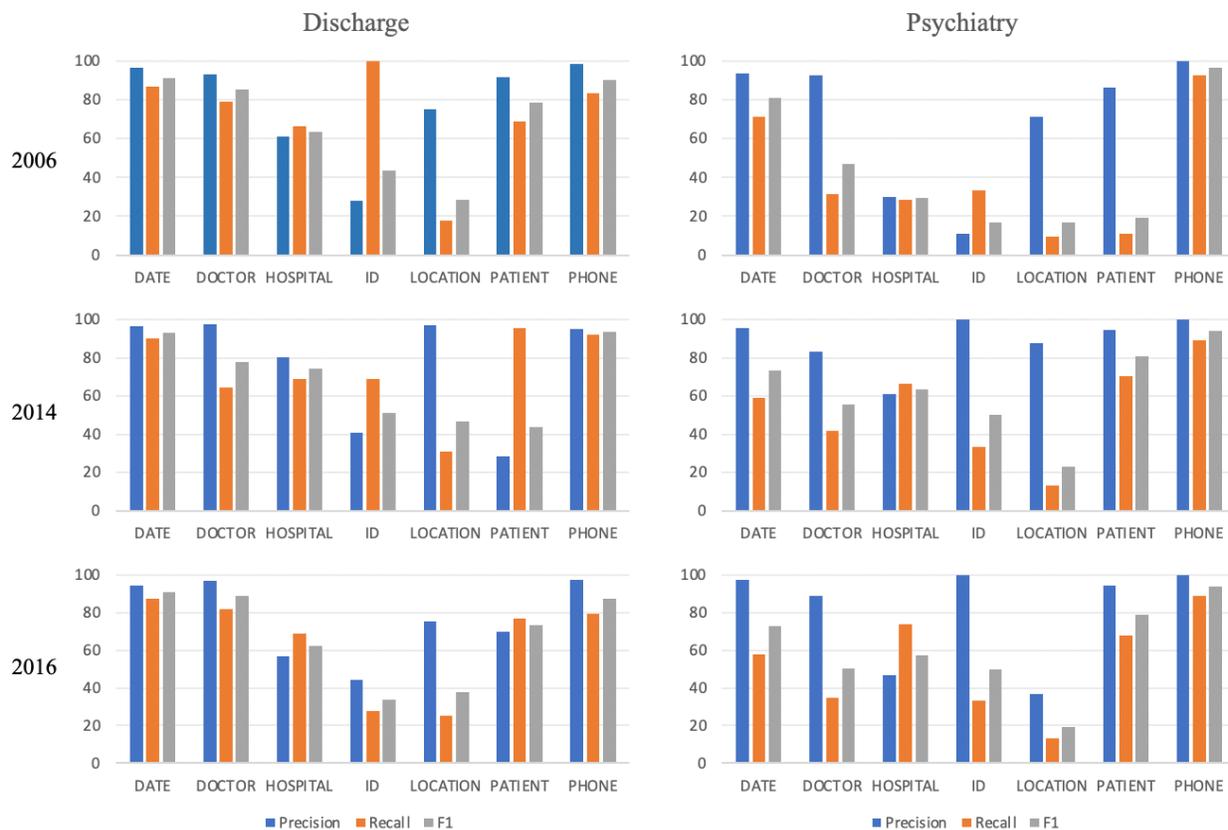

**Figure B.1.** Performance of individual Partners HealthCare-trained models on UW discharge and psychiatry test dataset notes by PHI type using NeuroNER architecture.

|  |  | Source | |
|---|---|---|---|
|  |  | UW (no pretraining) | UW w/ Partners HealthCare pretrained |
| **Target (UW)** | Age | 0/0/0 | 100.0/100.0/100.0 |
|  | Date | 98.22/97.7/98.0 | 98.6/98.1/98.4 |
|  | Doctor | 85.9/87.2/86.6 | 81.3/91.7/86.2 |
|  | Hospital | 79.5/84.2/81.8 | 76.1/85.7/80.6 |
|  | ID | 65.1/53.3/58.7 | 69.1/93.2/79.3 |
|  | Location | 85.7/76.1/80.6 | 85.9/77.9/81.7 |
|  | Patient | 86.1/87.1/86.6 | 90.9/83.4/87.0 |
|  | Phone | 74.5/90.6/81.8 | 93.4/85.2/89.1 |

**Table B.2.** Transferability from combined Partners HealthCare to UW by PHI type using NeuroNER architecture (%, P/R/$F_1$). 'ID' and 'Phone' PHI types show the greatest $F_1$-score improvement when fine-tuning (79.3 versus 58.7 and 89.1 versus 81.8), while other PHI types show only slight changes.

| Architecture | Training Strategy | Source | Target | Performance |
|---|---|---|---|---|
| NeuroNER | In-domain | UW | UW | 90.5/91.8/91.1 |
|  | Out-domain | 2014 | UW | 78.3/76.6//77.4 |
|  | Concurrent | 2014+UW | UW | 90.4/91.7/91.0 |
|  | Fine-tuning | 2014→UW | UW | **90.8/92.2/91.5** |
|  | In-domain | 2014 | 2014 | 94.8/92.7/93.8 |
|  | Out-domain | UW | 2014 | 86.0/73.8/79.4 |
|  | Concurrent | UW+2014 | 2014 | 95.3/92.4/93.8 |
|  | Fine-tuning | UW→2014 | 2014 | **95.4/93.2/94.3** |

**Table B.3** Transferability between i2b2 2014 dataset and UW (%, P/R/$F_1$). Best performance for each target set is in bold. The differences in $F_1$-scores between the best and second best-performing training strategies were statistically significant ($p \leq 0.05$) using approximate randomization [29] and 10,000 shuffles.